\documentclass[runningheads,orivec]{llncs}
\usepackage[T1]{fontenc}
\usepackage[utf8]{inputenc}
\usepackage{lmodern}
\usepackage{amsmath,amssymb}
\usepackage{graphicx}
\usepackage{booktabs,tabularx,array}
\usepackage{listings}
\usepackage{xcolor}
\usepackage{microtype}
\usepackage{placeins}
\usepackage{xurl}
\usepackage{hyperref}
\hypersetup{colorlinks=true,linkcolor=blue,citecolor=blue,urlcolor=blue,
  pdftitle={Vague2Detect: Handling Ambiguous Prompts in Knowledge-Based Open-World Detection},
  pdfauthor={Ibrohimjon Muminov and Jihie Kim}}
\graphicspath{{figures/}}
\newcommand{\exampleheading}[1]{\par\medskip\noindent\textbf{#1}\par\nobreak\smallskip}
\begin{document}
\title{Vague2Detect: Handling Ambiguous Prompts in Knowledge-Based Open-World Detection}
\titlerunning{Vague2Detect}
\author{Ibrohimjon Muminov \and Jihie Kim}
\authorrunning{Ibrohimjon Muminov and Jihie Kim}
\institute{Department of Computer Science and AI, Dongguk University, Seoul, South Korea\\
\email{2024126770@dongguk.edu, jihie.kim@dgu.edu}}
\maketitle
\begin{abstract}
Real-world detectors must often interpret functional or ambiguous prompts, yet conventional models such as YOLO remain restricted to fixed class lists. Even open-vocabulary models like YOLO-World frequently misalign vague language with the intended objects. Building on our prior work Commonsense-Guided Open-World Object Detection Using LLMs and Visual-Semantic Matching, we address YOLO-World's limitations in grounding task-driven queries. 

We propose Vague2Detect, a hybrid pipeline in which a fine-tuned Sentence-BERT retrieves candidates from a structured household Knowledge Base (KB), and YOLO-World verifies their presence in the image. For prompts outside the KB, a large language model (GPT-3.5-turbo) generates candidate descriptions, dynamically expanding the KB to cover novel concepts. 

On a benchmark of household scenes (custom images and Open Images V7 subset), YOLO-World alone achieves only 32\% Vague Prompt Success Rate (VPSR)---the ability to map ambiguous queries to correct detections. In contrast, Vague2Detect improves performance to 61\% VPSR with high precision, and up to 85\% when augmented with GPT fallback.

\noindent Code: \url{https://github.com/ibrohimgets/Vague2Detect}.
\end{abstract}

\section{Introduction}

Object detection has traditionally focused on recognizing instances of pre-defined categories (e.g., detecting ``sink'' or ``soap'' when those class labels are known). Modern detectors such as the YOLO family are extremely efficient and accurate for a fixed set of categories \cite{ref11}. However, their reliance on training-time category definitions limits applicability in open-world scenarios. Recent open-vocabulary methods such as YOLO-World \cite{ref3} extend coverage by combining vision and language, enabling detection beyond closed training sets. Yet, these models still expect queries to be explicit object labels or closely related textual variants.

A key gap remains in handling vague prompts, where users describe objects by function or affordance rather than by name (e.g., ``something to keep leftovers'' for a container, ``something to drink from'' for a cup). Standard benchmarks such as COCO \cite{ref11} emphasize category labels and do not evaluate functional language. As a result, models can achieve high mAP on COCO yet fail when queries are phrased indirectly. In our preliminary tests, baseline YOLO-World frequently misaligned vague prompts with visual categories---a failure mode not captured by conventional evaluation.

We propose Vague2Detect, a hybrid knowledge-based detection pipeline that combines symbolic knowledge and neural vision--language models to resolve such cases. The core idea is to interpret user intent using a household-object Knowledge Base (KB) with rich textual descriptions---bootstrapped with GPT for scalability---and then verify candidates with an object detector. Concretely, our system integrates: (i) a curated KB of 100+ household objects with appearance and usage descriptions, (ii) a fine-tuned Sentence-BERT model \cite{ref17} to map vague queries to KB categories, (iii) YOLO-World for visual verification, and (iv) a GPT-based fallback to generate candidate descriptions for out-of-KB queries.

This work extends our prior framework, Commonsense-Guided Open-World Object Detection Using LLMs and Visual-Semantic Matching \cite{ref13}, which showed that GPT-generated descriptions improve recall on novel queries. Unlike that system---which succeeded only when ambiguous prompts mapped to objects already in the KB---Vague2Detect tackles the harder setting of affordance-driven prompts, introduces semantic grounding via fine-tuned SBERT, and contributes a new benchmark and metric for evaluating success on vague queries.

\subsubsection*{Contributions}
The main contributions of this work are as follows:

\begin{itemize}
\item We propose Vague2Detect, a hybrid pipeline combining an LLM-bootstrapped knowledge base, fine-tuned SBERT, and YOLO-World to resolve vague prompts.
\item We fine-tune SBERT for functional grounding and introduce a GPT-based fallback for out-of-KB queries.
\item We build a benchmark from Open Images and our own household environment data, and define the Vague Prompt Success Rate (VPSR).
\item We conduct extensive evaluation and ablations, showing the complementary benefits of KB, visual verification, and LLM.
\end{itemize}

\section{Related Work}

Our prior work, Commonsense-Guided Open-World Object Detection Using LLMs and Visual-Semantic Matching \cite{ref13}, integrated YOLOv8, CLIP, LLaVA, GPT-4, and a BERT-based knowledge base into a single pipeline. In that framework, GPT-4 was employed offline to generate descriptions for over 100 household objects, resulting in a static KB. While this approach was effective for affordance-driven prompts that were already represented in the KB, it suffered from several important limitations: (i) the KB could not be expanded during inference, restricting adaptability to unseen concepts, (ii) the unfine-tuned BERT encoder yielded poor retrieval performance for ambiguous or functional queries, and (iii) running multiple large-scale models simultaneously (YOLOv8, CLIP, LLaVA, and GPT-4) introduced heavy computational overhead and made the system less practical for real-time or resource-constrained settings.

Vague2Detect addresses these shortcomings by fine-tuning SBERT \cite{ref17} for more robust and efficient prompt grounding, adopting YOLO-World \cite{ref3} as a unified open-vocabulary detector, and employing a single LLM both to bootstrap the KB and to act as a lightweight, dynamic fallback. This streamlined design eliminates the need to run several heavyweight models in parallel, while enabling continual KB expansion and efficient handling of vague prompts. As a result, the proposed system is both lighter and more adaptive compared to our earlier framework.

\subsection{Open-Vocabulary Object Detection}

Closed-set detectors rely on fixed training categories, limiting their ability to generalize. Open-vocabulary detection (OVD) addresses this by aligning vision and language. CLIP \cite{ref16} demonstrated zero-shot classification, inspiring extensions such as ViLD \cite{ref5}, Detic \cite{ref21}, and GLIP \cite{ref9}. YOLO-World \cite{ref3} further integrated vision--language pretraining into the YOLO framework, achieving strong zero-shot detection. However, these methods still assume explicit labels or near-synonyms, leaving functional or vague prompts unresolved. Our work fills this gap by introducing semantic grounding via a knowledge base and fine-tuned SBERT \cite{ref17}.

\subsection{Task-Driven and Affordance-Based Detection}

Users often describe objects by function rather than name. Sawatzky et al. \cite{ref18} introduced task-driven detection, TOIST \cite{ref10} exploited task context, and CoT-Det \cite{ref20} applied chain-of-thought prompting. These approaches rely on transient reasoning or context alone. By contrast, Vague2Detect employs a persistent KB and fine-tuned SBERT \cite{ref17}, enabling reusable and efficient grounding of affordance-driven prompts.

\subsection{Knowledge Bases in Vision and Language}

Knowledge resources such as ConceptNet \cite{ref19} and Visual Genome \cite{ref7} have long been used to enrich visual models with relational and commonsense context. However, these resources are static: they provide broad coverage but cannot easily adapt to new or rare concepts that arise in real-world queries. More recent approaches bypass static resources by querying large language models (LLMs) directly, gaining scalability but often sacrificing interpretability and control.

Our design seeks a middle ground. We employ a curated household Knowledge Base (KB) to ensure efficiency, transparency, and reproducibility, while reserving LLM fallback only for rare or unseen prompts. This allows the system to dynamically expand its vocabulary without over-reliance on costly LLM queries. Finally, YOLO-World \cite{ref3} provides visual verification, ensuring that newly added concepts are grounded in actual detections and reducing the risk of hallucinations.

\subsection{Sentence-BERT for Semantic Similarity}

Sentence-BERT (SBERT) \cite{ref17} encodes sentences into dense embeddings that can be compared using cosine similarity, enabling efficient retrieval in semantic search tasks. This makes it particularly well-suited for mapping vague or functional prompts (e.g., ``something to dry hands'') to the closest object descriptions in a Knowledge Base. Unlike traditional bag-of-words or TF-IDF approaches, SBERT captures contextual meaning, allowing the model to relate queries like ``I need to write a note'' to objects such as ``pen'' or ``pencil.''

For our baseline, we adopt the BGE model BAAI/bge-base-en-v1.5, a strong general-purpose retriever that has demonstrated competitive performance on large-scale semantic similarity benchmarks. However, since it is trained on broad web and QA datasets, it often struggles with affordance-driven or household-specific prompts, leading to mismatches (e.g., aligning ``something to cut food'' with ``plate'' instead of ``knife'').

To address this, we fine-tune SBERT on a curated set of 500 ambiguous prompt--object pairs. Importantly, these prompts are not direct copies of the KB entries but instead express the same target objects through varied descriptions and affordance-based usages (e.g., ``something to eat noodles with'' $\rightarrow$ ``fork''; ``a large cold storage appliance'' $\rightarrow$ ``refrigerator''). This diversification exposes the model to functional and descriptive variations that go beyond the static KB definitions. Fine-tuning in this way adapts the embedding space to household affordances, significantly improving grounding accuracy over the baseline BGE embeddings. Empirically, this reduces semantic drift and ensures that retrieved candidates align more closely with the intended task-driven meaning.

\subsection{Large Language Models in Vision Tasks}

Large language models (LLMs) such as GPT-3 \cite{ref2} and GPT-4 \cite{ref14} have powered recent multimodal systems including Flamingo \cite{ref1}, PaLM-E \cite{ref4}, and LLaVA \cite{ref12}, demonstrating impressive zero-shot reasoning across language and vision. CoT-Det \cite{ref20} applied chain-of-thought prompting for affordance reasoning, showing that LLMs can bridge functional intent and object recognition.

In our pipeline, GPT plays a more grounded and controlled role. Instead of directly generating detections, GPT is invoked only as a fallback: when SBERT retrieval fails to match a prompt to the Knowledge Base (KB), GPT produces candidate object descriptions. These candidates are then validated by YOLO-World \cite{ref3} before being added to the KB. This ensures controlled expansion---scalable enough to capture unseen objects, but with reduced risk of hallucinations and limited reliance on costly LLM queries.

In summary, our work unites open-vocabulary detection, affordance reasoning, and knowledge-based grounding. Unlike prior approaches, it (i) dynamically expands its KB at inference time with verified GPT suggestions, (ii) fine-tunes SBERT \cite{ref17} for robust vague prompt retrieval, and (iii) introduces a dedicated benchmark and metric (VPSR) to evaluate system performance under functional queries.

\section{Methodology}

\begin{figure}[tbp]
\centering
\includegraphics[width=\textwidth]{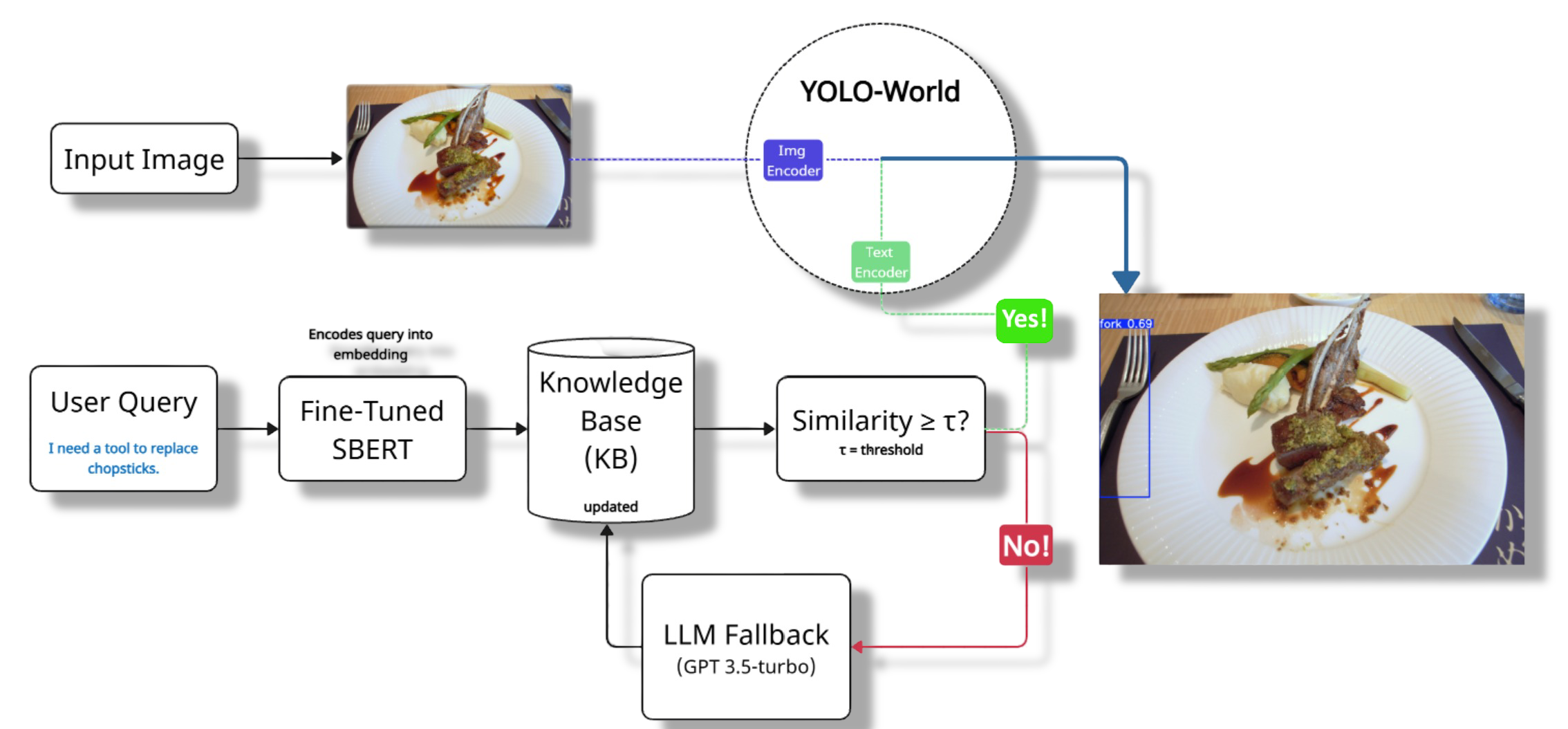}
\caption{Architecture of the proposed Vague2Detect pipeline. Given a vague query such as ``I need a tool to replace chopsticks,'' the system encodes the prompt with a fine-tuned SBERT model and retrieves the closest entry from the Knowledge Base (KB). If the retrieved object (e.g., fork) is above the similarity threshold $\tau$, its label is passed to YOLO-World for visual grounding. If no suitable KB match is found (for instance, if fork is not present in the KB), a GPT-based fallback generates a new candidate description for fork, which is then added to the KB and verified by YOLO-World before being accepted. This hybrid design enables the system to handle direct labels, functional queries, and novel objects while ensuring that all detections remain visually grounded.}
\label{fig:architecture}
\end{figure}

The goal of our system is to bridge the gap between vague, affordance-driven language and concrete visual object detection. To achieve this, we integrate structured knowledge with neural embeddings in a unified pipeline (Figure~\ref{fig:architecture}). A user query is first encoded with a fine-tuned SBERT model, which retrieves the most semantically relevant object from the Knowledge Base (KB). If the similarity score exceeds a threshold $\tau$, the candidate label is passed to YOLO-World \cite{ref3} for visual verification. This ensures that the system not only interprets the query linguistically but also confirms its visual presence in the scene.

If no KB entry meets the similarity threshold, the query is deemed out-of-coverage. In this case, a GPT-based fallback \cite{ref2,ref14} generates new candidate descriptions (visual description + usage), which are then added to the KB and validated by YOLO-World before being accepted. In this way, the system expands its knowledge dynamically while remaining grounded in visual evidence.

Overall, this hybrid design explicitly addresses three challenges: (i) recognizing explicit object labels, (ii) resolving functional or ambiguous prompts, and (iii) adapting to unseen concepts at inference time. By combining symbolic memory (KB), semantic retrieval (SBERT), and visual verification (YOLO-World), the proposed Vague2Detect pipeline offers a scalable and reliable solution for open-world object detection.

\subsection{Knowledge Base Construction}

The Knowledge Base (KB) serves as the semantic memory of our system. We focus on household objects (e.g., sink, soap, refrigerator, microwave, toothbrush), as our target domain is indoor environments and everyday tasks where functional prompts are common. To bootstrap the KB, we used GPT-5 \cite{ref15} to generate textual entries for over 100 objects. Each entry contains two complementary fields:

\begin{itemize}
\item Visual Description: a short account of physical appearance (e.g., ``A sharp metal blade with a handle.''),
\item Usage Description: a functional explanation (e.g., ``Used for cutting and slicing food.'').
\end{itemize}

This structured representation allows SBERT to align both perceptual and affordance-driven queries with object classes.

\noindent\textbf{Ambiguous Prompt Corpus.} To fine-tune SBERT (Sec.~\ref{sec:encoder}), we constructed a dataset of $\sim$500 prompt--object pairs. For each of the 100+ KB objects, we created five vague but semantically related prompts that intentionally differed from the KB phrasing (e.g., ``What tool slices vegetables?'' $\rightarrow$ knife). These alternative expressions capture functional variation and ensure the model learns to generalize beyond the canonical KB entries. The corpus was used only for fine-tuning and then discarded to prevent data leakage. At test time, the KB consists solely of the canonical GPT-generated entries, with embeddings pre-computed for efficient similarity search.

\subsection{Semantic Similarity Encoder (SBERT)}\label{sec:encoder}

To interpret a user's free-form prompt, we require a mechanism to map text queries to relevant objects. We employ Sentence-BERT (SBERT) \cite{ref17} as our text encoder. SBERT outputs dense embeddings for sentences such that semantically similar sentences lie close in vector space. This allows us to compare a user query with KB descriptions.

\noindent\textbf{Baseline.} As a baseline we use the BGE model (BAAI/bge-base-en-v1.5), which outputs 768-dimensional embeddings trained for universal retrieval. At runtime, the query $q$ and each KB description $d_o$ are encoded into dense vectors $\operatorname{enc}(q),\operatorname{enc}(d_o)\in\mathbb{R}^{768}$. We compute cosine similarity as

\begin{equation}
S(q,d_o)=\frac{\operatorname{enc}(q)\cdot\operatorname{enc}(d_o)}
{\lVert\operatorname{enc}(q)\rVert\,\lVert\operatorname{enc}(d_o)\rVert}.
\label{eq:similarity}
\end{equation}
The predicted object $o^{*}$ is then chosen as

\begin{equation}
o^{*}=\underset{o}{\arg\max}\ S(q,d_o).
\label{eq:prediction}
\end{equation}
In words, the KB entry whose embedding is most similar to the query embedding is selected.

\noindent\textbf{Fine-Tuning.} The vanilla model often misaligned functional prompts. For example, given ``I want to wash my hands,'' it might incorrectly select towel instead of sink. To improve grounding, we fine-tuned SBERT on the Ambiguous Prompt Corpus. Each query was mapped to one of the canonical KB objects. We initialized the encoder from bge-base-en-v1.5, added a classification head with output units equal to the KB size, and trained with cross-entropy loss using Hugging Face's Trainer API. Training used batch size 2, learning rate $2\times10^{-5}$, weight decay 0.01, and 10 epochs with mixed precision (fp16). Model selection was based on accuracy and macro-F1 on a held-out 20\% split.

\noindent\textbf{Inference.} At test time, SBERT encodes the query and compares it against all KB entries using cosine similarity. If the maximum score

\begin{equation}
S_{\max}=\max_o S(q,d_o)
\label{eq:maximum}
\end{equation}
exceeds a threshold $\tau$ (set to 0.6), the corresponding object $o^{*}$ is accepted as the predicted class. Otherwise, the query is deemed ambiguous and the fallback module is triggered.

\subsection{Visual Verification with YOLO-World}\label{sec:verification}

SBERT provides semantic grounding but cannot confirm whether the predicted object actually appears in the image. To ensure correctness, we use YOLO-World \cite{ref3}, an open-vocabulary extension of YOLOv8 with a CLIP-based text encoder \cite{ref16}. We use it directly without fine-tuning, relying on its ability to detect arbitrary text-specified classes.

If SBERT confidence is high ($S_{\max}\geq\tau$), only the top candidate $o_{\max}$ is checked. If confidence is low ($S_{\max}<\tau$), the top-$k$ candidates (e.g., $k=3$) are checked. YOLO-World accepts detections above 0.25 confidence: if at least one bounding box is found, the highest-confidence result is returned; otherwise, LLM fallback is triggered (Sec.~\ref{sec:fallback}).

This step prevents semantic errors. For instance, SBERT may suggest paintbrush for ``something to write with,'' but YOLO-World rejects it if absent, allowing fallback to propose pen or pencil. Since YOLO-World runs in real time ($\sim$50 FPS on modern GPUs) and only checks a few candidates, the computational cost is negligible compared to GPT calls.

\subsection{GPT-Based Knowledge Generation Fallback}\label{sec:fallback}
Even with fine-tuned SBERT retrieval and YOLO-World verification, some prompts may not map to any existing KB entry (e.g., ``something to listen to music with'' when no speaker entry exists; see the Appendix for a qualitative example). To handle such out-of-KB cases, we employ GPT-3.5-Turbo \cite{ref2} as a dynamic fallback mechanism.

\noindent\textbf{Prompting.} At runtime, the vague user query $q$ is first checked against the KB. If no sufficiently similar entry is found, the system activates the GPT fallback with explicit instructions: (i) infer 1--3 concrete object candidates, (ii) provide for each a short visual\_description and usage, and (iii) return the result in JSON format. The new entry is then added to the KB for future queries, ensuring continual growth of the knowledge base.

A typical example is shown below:

\begin{table}[tbp]
\caption{System log for the vague prompt ``something to listen to music with.''}
\label{tab:system-log}
\centering\small
\begin{tabularx}{\linewidth}{@{}lX@{}}
\toprule
Step & Output\\
\midrule
Prompt & Something to listen to music with\\
KB match & None (no similar object found)\\
Fallback & GPT generates new entry: ``speaker''\\
Action & Visual description and usage added to KB\\
\bottomrule
\end{tabularx}
\end{table}

\begin{lstlisting}
"speaker": {
  "visual_description": "A box-shaped device with a front cover and simple buttons or knobs",
  "usage": "Plays sound so people can listen to music, voices, or other audio"
}
\end{lstlisting}

\noindent\textbf{Integration.} The JSON response is parsed and merged into the KB, permanently expanding it for future queries. The system validates LLM outputs, rejecting malformed or non-JSON responses.

\noindent\textbf{Verification.} Newly generated entries are not accepted blindly. Each candidate is passed to YOLO-World for visual verification (Sec.~\ref{sec:verification}). Only objects detected in the current image are retained, preventing hallucinated KB expansion. In practice, this design allows the KB to grow dynamically over time, while keeping expansion grounded in the visual evidence of the current scene.

\section{Benchmark Setup}
\subsection{Dataset Construction}
We constructed a benchmark of $\sim$1000 images covering household object classes. The dataset combines samples from Open Images \cite{ref8} with photographs collected in our own indoor environments to better reflect everyday household settings. Each entry includes the raw image, a target object label, and a bounding box annotation for evaluation.

The object classes are divided into two groups:

\begin{itemize}
\item In-KB objects: categories explicitly represented in the Knowledge Base (e.g., sink, microwave, toothbrush),
\item Out-of-KB objects: categories absent from the Knowledge Base (e.g., speaker, nutcracker).
\end{itemize}

This design allows us to test both aspects of the pipeline: (i) semantic retrieval and grounding for KB-covered objects, and (ii) GPT fallback for novel categories that require dynamic expansion. By mixing familiar and unfamiliar objects within realistic household scenes, the benchmark provides a controlled yet challenging setting for evaluating vague-prompt detection.

\subsection{Main Results}
We evaluate our approach on the Vague Prompt Benchmark, comparing against YOLO-World and ablated variants. As shown in Table~\ref{tab:results}, the YOLO-World baseline achieves only 32\% vague prompt success rate (VPSR) and 29\% detection accuracy (IoU $\geq$ 0.5), highlighting its difficulty in resolving ambiguous queries.

Adding our fine-tuned SBERT module nearly doubles performance, reaching 61\% VPSR with a corresponding 61\% detection accuracy. Incorporating the GPT fallback provides the strongest results, with the full pipeline achieving 85\% VPSR and 83\% detection accuracy.

These results demonstrate that (i) SBERT fine-tuning is critical for grounding vague prompts, and (ii) GPT fallback extends coverage to unseen concepts while maintaining strong visual verification.

\begin{table}[tbp]
\caption{Results on the Vague Prompt Benchmark. VPSR = Vague Prompt Success Rate; Det.Acc = detection accuracy (IoU $\geq 0.5$).}
\label{tab:results}
\centering\small
\begin{tabular}{@{}lrr@{}}
\toprule
Configuration & VPSR (\%) & Det.Acc (\%)\\
\midrule
YOLO-World baseline (no KB) & 32 & 29\\
SBERT (fine-tuned) + YOLO-W & 61 & 61\\
Full pipeline (with GPT) & 85 & 83\\
\bottomrule
\end{tabular}
\end{table}

\subsection{Qualitative Results}
We present three representative examples that highlight the strengths and limitations of our pipeline:

\begin{enumerate}
\item Successful grounding within the KB. Fine-tuned SBERT correctly maps the vague prompt ``I want to wash my hands'' to the object class sink, which is then visually verified by YOLO-World.
\item Generalization to unseen concepts. When given prompts outside the KB, GPT fallback enables controlled expansion. For example, the query ``I want to play some music'' introduces the unseen class speaker, which GPT generates, adds to the KB, and YOLO-World successfully detects.
\item Failure modes. Common errors include fine-grained distinctions (e.g., confusing salt with pepper), small or occluded objects, and prompts with multiple valid answers. For instance, given ``I need something to cut with,'' the YOLO-World baseline fails to respond, vanilla SBERT confuses knife with peeler, while fine-tuning yields the correct mapping to knife.
\end{enumerate}

These cases demonstrate how our system (i) reliably grounds vague prompts within the KB, (ii) adapts to unseen objects via LLM fallback, and (iii) reveals remaining challenges in fine-grained recognition and ambiguity handling.

\section{Experimental Setup}

\subsection{Implementation Details}

All experiments were conducted on a single NVIDIA RTX 3090 GPU with 24GB memory. We fine-tuned Sentence-BERT on our ambiguous prompt--object corpus using the Adam optimizer \cite{ref6} (learning rate 1e-5, batch size 32) for 5 epochs. Embeddings for all Knowledge Base (KB) entries were pre-computed to enable efficient similarity search at inference time. YOLO-World \cite{ref3} was used for detection with a confidence threshold of 0.25 and non-maximum suppression applied by default.

\subsection{Dataset}

We constructed a benchmark of approximately 1000 images spanning 60 household-object classes. The dataset combines samples from Open Images \cite{ref8} with photographs collected in real indoor environments to better capture everyday household settings. Each entry consists of the raw image, a ground-truth object label, and a bounding box annotation.

The object classes include both in-KB categories, which are explicitly represented in our Knowledge Base (e.g., sink, table, microwave), and out-of-KB categories, which are absent from the KB and require LLM fallback for detection (e.g., speaker, air purifier). This split allows us to evaluate two complementary aspects of the pipeline: semantic retrieval and grounding for KB-covered objects, and dynamic expansion for novel concepts.

Overall, the benchmark provides a balanced mixture of common and rare household items, offering a controlled yet challenging testbed for vague-prompt detection.

\subsection{Evaluation Protocol}
At inference time, the system is provided with two inputs: (i) an image and (ii) a vague, affordance-driven prompt (e.g., ``something to cut fruits''). The task is to identify the most appropriate object and localize it in the scene.

Evaluation is based on two complementary metrics:

\begin{enumerate}
\item Class accuracy (Cls.Acc). A prediction is correct if the retrieved class matches the ground-truth label. We also report the Vague Prompt Success Rate (VPSR), which measures the proportion of ambiguous prompts that are successfully grounded to the correct object class.
\item Detection accuracy (Det.Acc). Beyond class grounding, the predicted bounding box must overlap with the ground-truth annotation by intersection-over-union (IoU) $\geq$ 0.5. This evaluates whether the system can not only interpret the prompt semantically but also localize the correct object visually.
\end{enumerate}

Together, Cls.Acc and Det.Acc capture both linguistic understanding and visual grounding, providing a comprehensive measure of vague-prompt object detection performance.

\subsection{Baselines}
We compare against three configurations, summarized in Table~\ref{tab:baselines}. The YOLO-World baseline applies detection directly without any semantic grounding. The SBERT + YOLO configuration introduces a fine-tuned semantic encoder that maps vague prompts to KB entries, verified by YOLO-World. Finally, the Full pipeline includes a GPT fallback for out-of-KB queries, allowing continual KB expansion and greater coverage.

\begin{table}[tbp]
\caption{Baseline configurations used for evaluation.}
\label{tab:baselines}
\centering\small
\begin{tabularx}{\linewidth}{@{}p{0.27\linewidth}X@{}}
\toprule
Baseline & Description\\
\midrule
YOLO-World & Applied directly without KB or semantic grounding.\\
SBERT + YOLO-W & Fine-tuned SBERT retrieves candidate class; YOLO verifies.\\
Full pipeline & Includes LLM fallback for out-of-KB queries, enabling KB expansion.\\
\bottomrule
\end{tabularx}
\end{table}

\subsection{Knowledge Base Expansion}
Beyond baseline comparisons, we also examine how the Knowledge Base (KB) adapts under different configurations. In the static baseline, the KB remains fixed at 100 predefined household objects. In contrast, our full pipeline is dynamic: when a user provides a rare or unconventional prompt (e.g., ``nutcracker'' or ``speaker''), the system falls back to GPT to generate and validate new entries. This mechanism ensures coverage for rare cases, while the majority of queries are successfully resolved within the KB itself, keeping GPT usage minimal.

\FloatBarrier
\section{Conclusion}
We introduced Vague2Detect, a hybrid pipeline for vague-prompt object detection that combines four components: (i) a curated Knowledge Base for semantic memory, (ii) a fine-tuned SBERT encoder for mapping functional language to object classes, (iii) YOLO-World for visual verification, and (iv) a GPT-based fallback to expand coverage for unseen categories.

On a benchmark of household scenes constructed from Open Images and our own photographs, Vague2Detect substantially outperformed strong baselines, reaching 85\% vague prompt success rate (VPSR) and 83\% detection accuracy (Det.Acc). Ablation studies confirmed the contribution of each module: SBERT improved functional grounding, YOLO-World ensured visual verification, and GPT fallback enabled dynamic knowledge expansion.

\noindent\textbf{Limitations.} The current system is constrained by (a) reliance on a relatively small Knowledge Base (approximately 100 objects), which may limit coverage in diverse domains, (b) occasional errors in fine-grained or visually similar classes (e.g., salt vs. pepper), and (c) dependence on GPT fallback, which, while effective, adds computational cost.

\noindent\textbf{Future Work.} To address these limitations, we plan to (i) scale the Knowledge Base using semi-automatic generation across broader domains, (ii) explore lightweight adapters to reduce reliance on costly GPT fallback, (iii) incorporate spatio-temporal and commonsense reasoning for handling dynamic household activities, and (iv) extend the framework to robotic settings where affordance-driven perception directly informs action.

Overall, our modular design demonstrates that object detection can move beyond identifying what an object is toward reasoning about why it is needed, opening new directions for affordance-driven and task-oriented perception systems.

\subsubsection*{Acknowledgments}
This research was supported by the MSIT (Ministry of Science and ICT), Korea, under the ITRC (Information Technology Research Center) support program (IITP-2025-RS-2020-II201789), and the Artificial Intelligence Convergence Innovation Human Resources Development (IITP-2025-RS-2023-00254592) supervised by the IITP (Institute for Information \& Communications Technology Planning \& Evaluation).

\clearpage
\raggedbottom
\appendix
\section*{Appendix}
\subsection*{Additional Results}
Here we provide additional qualitative examples that were omitted from the main paper due to space constraints. These cases further illustrate how our system handles functional or vague queries that do not directly mention the target object by name. For example, the system correctly interprets the vague prompt ``something to listen to music with'' as a speaker (generated through the fallback LLM since no suitable KB entry existed), ``I need something sharp to cut the fish'' as a knife (retrieved directly from the KB), and ``Where can I wash my dishes?'' as a sink (retrieved directly from the KB). These examples highlight the ability of our model to bridge the gap between everyday task-oriented language and concrete object categories, successfully grounding functional descriptions into visual detections.

\exampleheading{Example: Speaker}
\begin{lstlisting}[caption={System log for the vague prompt ``something to listen to music with.''}]
Prompt: Something to listen to music with
KB match: None (no similar object found)
Fallback activated -> GPT generates new entry: "speaker"
Visual description and usage added to KB

KB description: A box-shaped device that plays sound.
Detected object: speaker (confidence 0.98)
\end{lstlisting}
\begin{center}
\includegraphics[width=0.40\textwidth]{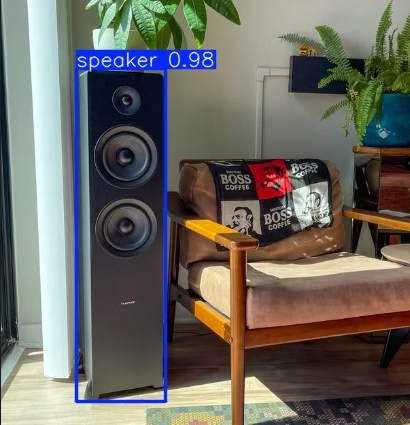}
\end{center}

\clearpage
\exampleheading{Example: Knife}
\begin{lstlisting}[caption={System log for the vague prompt ``I need something sharp to cut the fish.''}]
Prompt: I need something sharp to cut the fish
KB match: knife (found in KB, no fallback needed)

KB description: A sharp metal blade with a handle.
Detected object: knife (confidence 0.84)
\end{lstlisting}
\begin{center}
\includegraphics[width=0.52\textwidth]{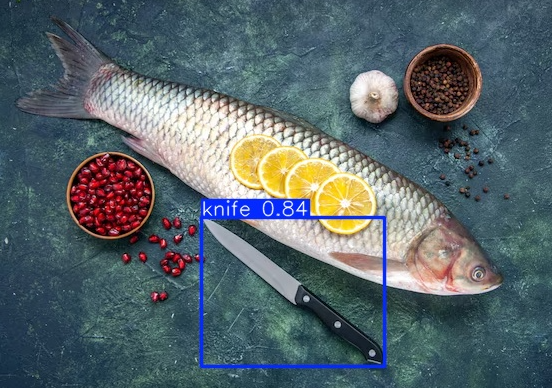}
\end{center}

\exampleheading{Example: Sink}
\begin{lstlisting}[caption={System log for the vague prompt ``Where can I wash my dishes?''}]
Prompt: Where can I wash my dishes?
KB match: sink (found in KB, no fallback needed)

KB description: A basin with a faucet, used for washing hands, dishes or food.
Detected object: sink (confidence 0.81)
\end{lstlisting}
\begin{center}
\includegraphics[width=0.52\textwidth]{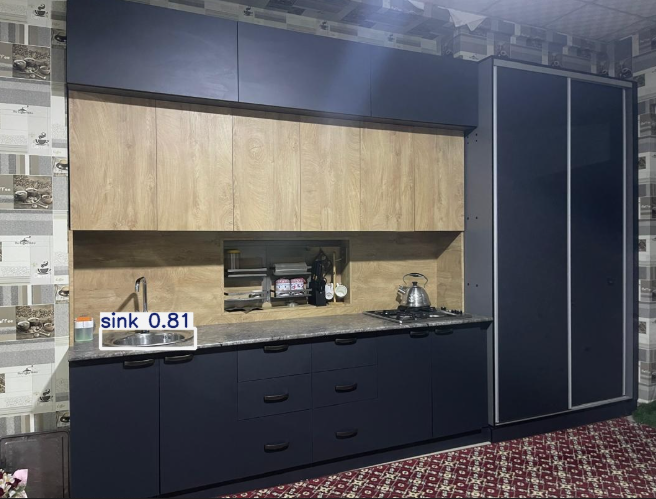}
\end{center}
\end{document}